\documentclass[a4paper,fleqn]{cas-dc}

\usepackage[authoryear]{natbib}
\usepackage{amssymb}
\usepackage{amsmath}
\usepackage{amsthm}
\usepackage{algorithm}
\usepackage{algpseudocode}
\usepackage{booktabs}
\usepackage{multirow}
\usepackage{graphicx}
\usepackage{microtype}
\usepackage{tikz}
\usepackage{url}

\usetikzlibrary{arrows.meta,calc,fit,positioning}

\definecolor{ptssanavy}{RGB}{22,45,84}
\definecolor{ptssateal}{RGB}{0,128,128}
\definecolor{ptssatealfill}{RGB}{230,247,245}
\definecolor{ptssared}{RGB}{181,72,62}
\definecolor{ptssaredfill}{RGB}{252,235,232}
\definecolor{ptssaamberfill}{RGB}{255,247,223}
\definecolor{ptssagrayfill}{RGB}{244,246,249}
\definecolor{ptssagraytext}{RGB}{82,91,105}

\newcommand{\ind}{\mathbb{1}}

\newcommand{\clip}{\operatorname{clip}}

\begin{document}

\let\WriteBookmarks\relax
\def\floatpagepagefraction{1}
\def\textpagefraction{.001}




\shorttitle{Parallel Time-Aligned Spiking Self-Attention}
\shortauthors{Xue et al.}

\title[mode=title]{Parallel Time-Aligned Spiking Self-Attention for Consistent Integer-Valued
Training and Spike-Driven Inference}

\author[1,2,5]{Peng Xue}
\credit{Conceptualization, Methodology, Software, Validation, Formal analysis, Investigation, Data curation, Visualization, Writing - Original draft}
\author[3]{Wei Fang}
\credit{Methodology, Validation, Writing - Review \& editing}
\author[3]{Kaiwei Che}
\credit{Validation, Investigation, Writing - Review \& editing}
\author[2]{Qingyan Meng}
\credit{Methodology, Writing - Review \& editing}
\author[2]{Zhengyu Ma}
\credit{Supervision, Writing - Review \& editing}
\author[2,3,4]{Yonghong Tian}
\cormark[1]
\ead{yhtian@pku.edu.cn}
\credit{Resources, Funding acquisition, Writing - Review \& editing}
\author[2]{Huihui Zhou}
\cormark[1]
\ead{zhouhh@pcl.ac.cn}
\credit{Supervision, Project administration, Funding acquisition, Writing - Review \& editing}

\cortext[1]{Corresponding authors}

\affiliation[1]{organization={Shenzhen Institute of Advanced Technology, Chinese Academy of Sciences},
country={China}}
\affiliation[2]{organization={Pengcheng Laboratory},
country={China}}
\affiliation[3]{organization={Shenzhen Graduate School, Peking University},
country={China}}
\affiliation[4]{organization={School of Computer Science, Peking University},
country={China}}
\affiliation[5]{organization={University of Chinese Academy of Sciences},
country={China}}

\begin{abstract}
Integer-valued leaky integrate-and-fire (I-LIF) neurons and spike firing approximation (SFA) reduce temporal training cost by representing spike trains as firing counts and normalized firing rates, respectively.
However, applying spiking self-attention (SSA) directly to these compressed query, key, and value representations introduces cross-time interactions that are absent during spike-driven inference.
We term this operator-level discrepancy Temporal Interaction Mismatch (TIM).
We propose Parallel Time-Aligned Spiking Self-Attention (PT-SSA), which reconstructs consecutive virtual spike slices from either I-LIF counts or SFA firing rates, computes attention only between time-aligned slices in parallel, and sums the per-step outputs.
To accommodate the reduced attention output scale under SFA, we further introduce Adaptive PT-SSA, which learns a positive per-block rescaling before the output SFA neuron to improve firing-level utilization.
Experiments on CIFAR-10, CIFAR-100, and ImageNet-1K show that the proposed methods substantially reduce train--inference mismatch.
On CIFAR-100 with I-LIF, PT-SSA reduces the mean Top-1 gap from 1.85 to 0.29 percentage points.
On ImageNet-1K, Adaptive PT-SSA reduces the Top-1 gap from 27.78 to 0.06 percentage points and achieves 74.53\% spike-driven Top-1 accuracy.
A Triton-fused PT-SSA training kernel retains a $2.91\times$ throughput advantage over recurrent LIF SSA.
\end{abstract}



\maketitle

\section{Introduction}
\label{sec:introduction}

Spiking Neural Networks (SNNs) process temporal information through stateful neuronal dynamics and communicate via sparse spikes \citep{maass1997networks}.
This spike-based communication enables energy-efficient, event-driven execution on neuromorphic processors such as Loihi \citep{davies2018loihi}.
Yet the non-differentiability of the discrete spike-generation function makes direct training challenging.
Surrogate-gradient methods provide a practical approximation by using a differentiable proxy for the spike derivative during backpropagation, thereby facilitating gradient-based training of SNNs \citep{neftci2019surrogate}.
Representative advances include SEW-ResNet for effective residual learning in deep SNNs \citep{fang2021deep}, Spikformer for Spiking Self-Attention (SSA) \citep{zhou2023spikformer}, and the Spike-Driven Transformer series for efficient spike-driven operators and general-purpose visual backbones \citep{yao2023spikedriven,yao2024spikedrivenv2,yao2025scaling}.
Other recent models, including QKFormer and MaxFormer, continue to refine spiking attention and feature representation \citep{zhou2024qkformer,fang2026spiking}.

Despite these architectural advances, explicit temporal training incurs substantial computation and memory overhead because it processes and stores spikes, membrane states, and gradients across time steps \citep{ma2026spikeaware}.
To mitigate this overhead, parallel spiking neurons remove reset-induced temporal recurrence, enabling parallel computation across time and substantially accelerating training \citep{fang2023parallel}.
Integer-valued LIF (I-LIF) instead represents a multi-step spike train as an integer activation during training and unfolds the integer into an equivalent spike sequence at inference, reducing temporal simulation cost while improving accuracy \citep{luo2024integer}.
Spike firing approximation (SFA) \citep{yao2025scaling} can be viewed as a normalized variant of I-LIF.
During training, it interprets the integer activation divided by the number of virtual time steps as the firing rate of the corresponding spike train.
At inference, the integer is expanded into a spike sequence, while the normalization factor is folded into the weights of the following stateless layer.

\begin{figure*}[t]
  \centering
  \includegraphics[width=\textwidth]{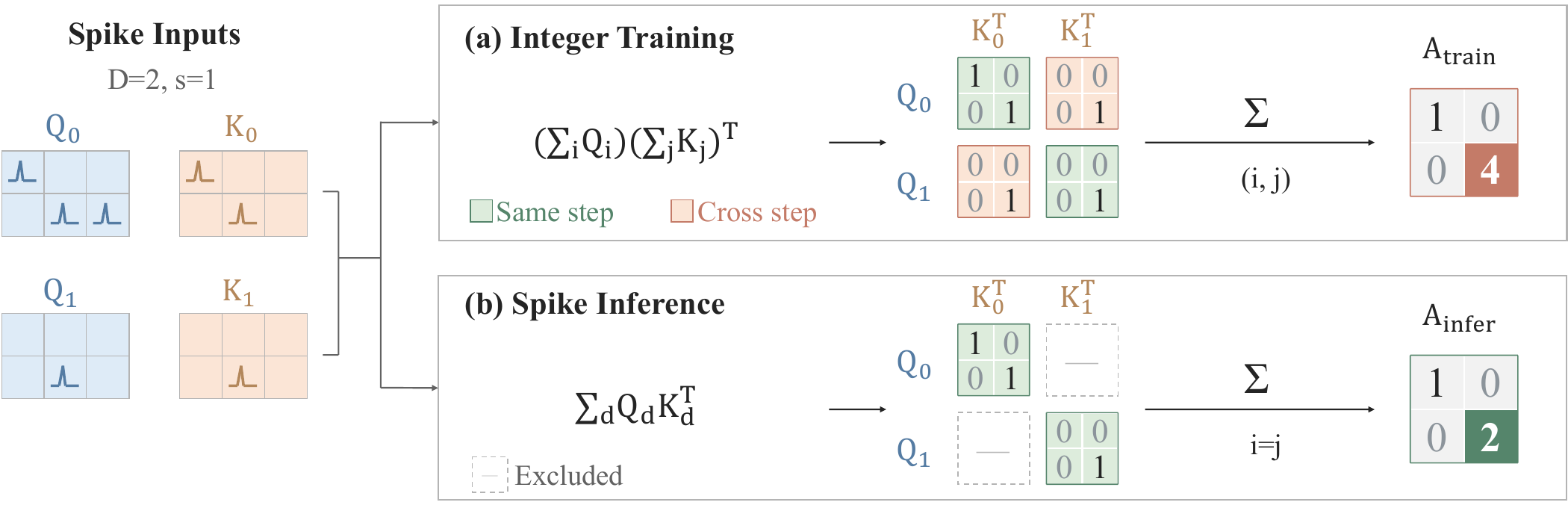}
  \caption{Temporal interaction mismatch in a two-step I-LIF example ($D=2$, $s=1$).
  (a) Multiplying compressed spike counts includes cross-time products and yields the highlighted value 4.
  (b) Summing same-step spike products yields 2 despite the same total spike counts.}
  \label{fig:train-inference-mismatch}
\end{figure*}

Although Spike-Driven Transformer V3 (SDT-V3) combines SFA with Efficient Spike-Driven Self-Attention (E-SDSA) and achieves state-of-the-art performance across vision tasks, its train--inference equivalence analysis is limited to linear operators \citep{yao2025scaling}.
We show that applying E-SDSA to SFA activations breaks operator-level equivalence between integer-valued training and temporal spike inference, a discrepancy in temporal interaction structure that we term Temporal Interaction Mismatch (TIM).
Specifically, applying E-SDSA to the SFA query ($Q$), key ($K$), and value ($V$) activations during training introduces cross-time interactions, whereas temporal spike inference retains only time-aligned interactions (Figure~\ref{fig:train-inference-mismatch}).
Our evaluation on ImageNet-1K shows that temporal spike inference incurs a Top-1 accuracy drop of more than 27 percentage points relative to integer-valued inference; reproducing the cross-time interactions at inference would instead require additional computation and storage for spatiotemporal attention on resource-constrained neuromorphic hardware such as Tianjic \citep{pei2019tianjic}.

To eliminate TIM, we formulate Parallel Time-Aligned Spiking Self-Attention (PT-SSA) for both I-LIF count and SFA firing-rate representations: PT-SSA reconstructs consecutive virtual spike slices from compact $Q$, $K$, and $V$ activations, computes attention only among time-aligned slices in parallel, and sums the per-step outputs.
Under the established SFA conversion assumptions and rules, this alignment preserves local equivalence between integer-valued training and spike-driven inference through the subsequent spiking neuron and linear projection \citep{yao2025scaling}.
Removing cross-time interactions also reduces the attention output scale, which can limit the utilization of firing levels in the output SFA neuron.
We therefore introduce Adaptive PT-SSA, which learns a positive per-block rescaling before this neuron to adapt its effective firing threshold to the aligned attention output.
Experiments validate the resulting end-to-end accuracy alignment: on ImageNet-1K, the Top-1 gap between integer and temporal spike inference is only 0.06 percentage points.

Our contributions are summarized as follows:
\begin{itemize}
  \item We identify and provide the first operator-level analysis of TIM in the SFA-based E-SDSA of SDT-V3, showing that it results in a 27.78-percentage-point drop in Top-1 accuracy on ImageNet-1K.
  \item We propose PT-SSA, which eliminates TIM for both I-LIF count and SFA firing-rate representations by reconstructing virtual spike slices and computing time-aligned attention in parallel.
  Building on this alignment, Adaptive PT-SSA learns a per-block rescaling to improve firing-level utilization under SFA.
  \item Experiments on CIFAR-10/100 and ImageNet-1K show that PT-SSA and Adaptive PT-SSA reduce the integer--spike accuracy gap, reducing the ImageNet-1K gap to 0.06 percentage points.
  The Triton-fused PT-SSA training kernel retains efficiency close to compressed integer SSA, with comparable peak memory and only 5.4\% lower throughput.
\end{itemize}

\section{Related Work}
\label{sec:related-work}

\subsection{Efficient SNN Training and Conversion}
\label{sec:related-neurons}

Conventional surrogate-gradient training optimizes SNNs through sequences of membrane states, making temporal simulation a major component of training cost \citep{neftci2019surrogate}.
One approach to accelerating direct SNN training uses parallel spiking neuron models that reformulate neuronal dynamics to reduce recurrent dependencies and enable computation across time \citep{fang2023parallel,xue2026multiplication}.
Fixed-point Parallel Training (FPT) also parallelizes temporal state computation, using a small number of fixed-point iterations to approximate the LIF state sequence \citep{feng2025efficient}.
A second approach reduces explicit temporal simulation during direct SNN training by representing multi-step spike trains as compact activations.
I-LIF uses integer firing counts, while SFA uses normalized firing rates \citep{luo2024integer,yao2025scaling}.
These representations avoid explicit simulation of the virtual time-step during training and are expanded into temporal spike sequences at inference.

ANN2SNN conversion constructs SNNs from pretrained ANNs by approximating analog activations with spike representations.
QCFS and Fast-SNN reduce conversion error through quantized activation design \citep{bu2022optimal,hu2023fast}, while offset-spike calibration addresses uneven spike generation \citep{hao2023bridging}.
Post-training quantization enables conversion calibration with a small dataset, using channel-wise firing thresholds to reduce conversion error without retraining \citep{sun2025posttraining}.
Recent work extends this paradigm through parallel spiking calculation \citep{hao2025faster} and differential coding, which transmits changes in rate information to support training-free conversion \citep{huang2025differential}.

\subsection{Spiking Transformers and Attention}
\label{sec:related-transformers}

Spiking Transformers have developed softmax-free attention and spike-compatible residual paths.
Spikformer introduced SSA with spike queries, keys, and values \citep{zhou2023spikformer}, while Spikingformer and the Spike-Driven Transformer series revised residual and attention computation for spike-driven visual backbones \citep{zhou2023spikingformer,yao2023spikedriven,yao2024spikedrivenv2}.
QKFormer develops hierarchical Q--K attention \citep{zhou2024qkformer}, and SDT-V3 combines E-SDSA with SFA to scale model capacity \citep{yao2025scaling}.
ESTSformer partitions attention heads into temporal and spatial groups to reduce the computational and storage costs of spatiotemporal attention \citep{lu2025estsformer}.
QSD-Transformer studies information distortion in quantized spike-driven attention and addresses it through an information-enhanced neuron and attention-distribution distillation \citep{qiu2025quantized}.
Its neuron produces multi-bit outputs during training while retaining spike-driven inference, connecting attention quantization with the choice of neuronal representation.

ANN-to-SNN conversion studies also address attention and nonlinear operators in Transformers.
Spatio-Temporal Approximation uses spatial approximations for nonlinear operations and temporal correction for spike multiplications in self-attention \citep{jiang2024spatiotemporal}.
SpikeZIP-TF instead constructs a Transformer conversion with exact ANN--SNN equivalence \citep{you2024spikezip}.
Our work instead studies train--inference consistency in SFA-based E-SDSA and addresses it with PT-SSA.

\subsection{Adaptive Neurons and Quantization}
\label{sec:related-quantization}

From a quantization perspective, SNN time steps and ANN activation bit widths offer related ways to control representational precision and resource demands \citep{shen2024conventional}.
Learned Step Size Quantization (LSQ) learns quantization step sizes and scales their gradients according to tensor size and quantization range \citep{esser2020lsq}.
Within SNNs, DIET-SNN learns leakage and firing thresholds to support low-latency inference \citep{rathi2023diet}, while DS-ATGO adapts thresholds and surrogate gradients to membrane-potential distributions \citep{jiang2026dsatgo}.
These approaches motivate adapting the activation scale or neuronal response to the statistics encountered during training.
Adaptive PT-SSA learns a rescaling before each attention-output SFA neuron to accommodate the output-scale change caused by excluding cross-time terms from the corresponding compressed attention computation.
This rescaling adjusts the effective firing threshold, while the alignment of temporal interactions is enforced by the attention operator itself.

\section{Preliminaries}
\label{sec:preliminaries}

Directly trained SNNs represent temporal information using stateful spiking neurons and sparse spike communication.
We review the neuron representations and spiking attention operator underlying Spike-Driven Transformer V3 (SDT-V3) \citep{yao2025scaling}.

\subsection{LIF, I-LIF, and SFA}
Discrete spiking neurons, including the leaky integrate-and-fire (LIF) neuron, can be described by charging, firing, and resetting dynamics \citep{fang2023spikingjelly}:
\begin{align}
  H[t] &= f\bigl(V[t-1],X[t]\bigr), \label{eq:lif-charge}\\
  S[t] &= \Theta\bigl(H[t]-V_{\mathrm{th}}\bigr), \label{eq:lif-fire}\\
  V[t] &=
  \begin{cases}
    H[t]\bigl(1-S[t]\bigr)+V_{\mathrm{reset}}S[t], & \text{hard reset},\\
    H[t]-V_{\mathrm{th}}S[t], & \text{soft reset},
  \end{cases}
  \label{eq:lif-reset}
\end{align}
where $X[t]$ is the input, $H[t]$ is the membrane potential after charging and before firing, $S[t]\in\{0,1\}$ is the output spike, and $V[t]$ is the post-reset membrane potential.
Here, $\Theta$ is the Heaviside function, $V_{\mathrm{th}}$ and $V_{\mathrm{reset}}$ are the firing threshold and reset potential, and $f$ implements leaky integration for LIF neurons.
The recurrence through $V[t-1]$ requires sequential simulation over time.

I-LIF replaces the spike-generation rule in Eq.~\eqref{eq:lif-fire} with the following integer-valued activation \citep{luo2024integer}:
\begin{equation}
  C[t]=\operatorname{clip}\bigl(\operatorname{round}(H[t]),0,D\bigr),
  \label{eq:ilif-fire}
\end{equation}
where membrane potential is expressed in threshold units and $D$ is both the maximum firing count and the number of virtual inference steps.
This multi-level activation reduces quantization error relative to a spike output without explicitly simulating the $D$ virtual steps during training.
At inference, each $C[t]$ is expanded into a length-$D$ spike sequence $\{S[t,d]\}_{d=0}^{D-1}$ satisfying $\sum_{d=0}^{D-1}S[t,d]=C[t]$; the resulting spike representation remains equivalent to the integer activation through a subsequent linear operator \citep{luo2024integer}.

For a tensor of such counts $\mathbf{C}_{X}$, SFA further normalizes the integer activation to the firing rate $\mathbf{C}_{X}/D$ during training and adopts the same spike-driven inference conversion \citep{yao2025scaling}.
We write the I-LIF count and SFA firing-rate activations in a common spike-count representation:
\begin{equation}
  \bar{\mathbf{X}} = \frac{\mathbf{C}_{X}}{s},
  \qquad
  s =
  \begin{cases}
    1, & \text{I-LIF count representation},\\
    D, & \text{SFA firing-rate representation},
  \end{cases}
  \label{eq:unified-representation}
\end{equation}
where $s=1$ gives the I-LIF count and $s=D$ gives the SFA firing rate.
Following the SFA conversion protocol, an integrate-and-fire neuron with soft reset starts from zero membrane potential, receives the count tensor $\mathbf{C}_{X}$ at the first virtual step, and receives zero input over the remaining $D-1$ steps \citep{yao2025scaling}.
With a unit firing threshold, the neuron subtracts one threshold unit after each spike and consequently produces the consecutive spike sequence
\begin{equation}
  \mathbf{X}_d=\ind[\mathbf{C}_{X}>d]\in\{0,1\},
  \qquad d=0,\ldots,D-1.
  \label{eq:consecutive-code}
\end{equation}
Thus, $\mathbf{C}_{X}=\sum_d\mathbf{X}_d$ and $\bar{\mathbf{X}}=s^{-1}\sum_d\mathbf{X}_d$.
For $T$ task-level steps, this conversion produces $TD$ inference steps in total.

\subsection{Spiking Self-Attention}
Spikformer introduced softmax-free Spiking Self-Attention (SSA), which performs spatial self-attention separately at each time step \citep{zhou2023spikformer}.
For one attention head, the output at time step $t$ has the generic form
\begin{equation}
  \mathbf{O}[t]
  = \alpha\bigl(\mathbf{Q}[t]\mathbf{K}[t]^{\mathsf T}\bigr)\mathbf{V}[t],
  \label{eq:step-attention}
\end{equation}
where $\alpha$ is the head scaling factor and $\mathbf{Q}[t]$, $\mathbf{K}[t]$, and $\mathbf{V}[t]$ are spike tensors with entries in $\{0,1\}$.
Representing $Q$, $K$, and $V$ as spikes reduces the dot products to logical AND and accumulation operations, while omitting softmax avoids dense floating-point normalization; together, these properties preserve sparse spike-based computation.

\subsection{Spike-Driven Transformer V3}
SDT-V3 combines SFA with Efficient Spike-Driven Self-Attention (E-SDSA) and reports state-of-the-art performance across multiple vision tasks \citep{yao2025scaling}.
E-SDSA generates $Q$, $K$, and $V$ with spiking neurons and applies the softmax-free $QK^{\mathsf T}V$ operator in Eq.~\eqref{eq:step-attention}.
During SFA training, E-SDSA receives the firing-rate tensors $\bar{\mathbf{Q}}=\mathbf{C}_Q/D$, $\bar{\mathbf{K}}=\mathbf{C}_K/D$, and $\bar{\mathbf{V}}=\mathbf{C}_V/D$ from Eq.~\eqref{eq:unified-representation}.
At inference, each count tensor $\mathbf{C}_X$ is converted into the corresponding $D$-step spike sequence $\{\mathbf{X}_d\}_{d=0}^{D-1}$ in Eq.~\eqref{eq:consecutive-code}, where $X\in\{Q,K,V\}$.
For this conversion, SFA proves $\mathbf{C}_X=\sum_d\mathbf{X}_d$ under the integrate-and-fire dynamics with soft reset described above and shows that $1/D$ can be absorbed into the weights when the subsequent operator is linear.
However, it does not analyze whether the firing-rate and spike-train representations remain equivalent through E-SDSA.

\section{Method}
\label{sec:method}

We first analyze the train--inference inconsistency introduced by applying SDT-V3's E-SDSA to its SFA firing-rate activations.
We then introduce Parallel Time-Aligned Spiking Self-Attention (PT-SSA), which restores time-aligned attention during parallel training.
Finally, we extend PT-SSA to Adaptive PT-SSA, which learns a rescaling of the attention output before the standard SFA neuron to accommodate the PT-SSA output scale under the normalized SFA representation.

For the analysis below, we treat SFA as a normalized special case of I-LIF and use Eq.~\eqref{eq:unified-representation}, with $s=1$ for I-LIF and $s=D$ for SFA.

\subsection{Analysis of Temporal Interaction Mismatch}
\label{sec:mismatch}

Let $\bar{\mathbf{Q}}$, $\bar{\mathbf{K}}$, and $\bar{\mathbf{V}}$ denote the training spike-count representations, and let $\mathbf{Q}_d$, $\mathbf{K}_d$, and $\mathbf{V}_d$ denote their corresponding spikes at virtual step $d$.
By Eqs.~\eqref{eq:unified-representation} and \eqref{eq:consecutive-code}, $\bar{\mathbf{X}}=s^{-1}\sum_d\mathbf{X}_d$ for $X\in\{Q,K,V\}$.
Applying attention directly to these training representations therefore yields
\begin{align}
  \mathbf{O}_{\mathrm{train}}
  &= \alpha
     \left(\bar{\mathbf{Q}}\bar{\mathbf{K}}^{\mathsf T}\right)
     \bar{\mathbf{V}} \nonumber\\
  &= \frac{\alpha}{s^3}\sum_{i,j,k}
     \left(\mathbf{Q}_i\mathbf{K}_j^{\mathsf T}\right)\mathbf{V}_k.
  \label{eq:compressed-attention}
\end{align}
At the same representation scale, temporal spike inference instead aggregates
\begin{equation}
  \mathbf{O}_{\mathrm{inference}}
  = \frac{\alpha}{s^3}\sum_{d=0}^{D-1}
    \left(\mathbf{Q}_d\mathbf{K}_d^{\mathsf T}\right)\mathbf{V}_d.
  \label{eq:reference-attention}
\end{equation}
Equation~\eqref{eq:compressed-attention} includes every triplet $(i,j,k)$, whereas Eq.~\eqref{eq:reference-attention} retains only $i=j=k$.
Their difference is therefore
\begin{equation}
  \mathbf{O}_{\mathrm{train}}-\mathbf{O}_{\mathrm{inference}}
  = \frac{\alpha}{s^3}\!\sum_{(i,j,k):\,\neg(i=j=k)}
    \left(\mathbf{Q}_i\mathbf{K}_j^{\mathsf T}\right)\mathbf{V}_k.
  \label{eq:cross-time-error}
\end{equation}
As illustrated in Figure~\ref{fig:train-inference-mismatch}, when $D=2$, the compressed query--key product contains the extra terms $Q_0K_1$ and $Q_1K_0$.
Thus, firing-count conservation is insufficient for operator-level equivalence through attention.
The two paths coincide only in degenerate cases such as $D=1$ or vanishing off-diagonal interactions.
We term this discrepancy in temporal interaction structure Temporal Interaction Mismatch (TIM).
It results from evaluating a time-local trilinear attention operator on temporally compressed firing-count or firing-rate representations.

\subsection{Parallel Time-Aligned Spiking Self-Attention}
\label{sec:ptssa}

\begin{figure*}[t]
  \centering
  \includegraphics[width=\textwidth]{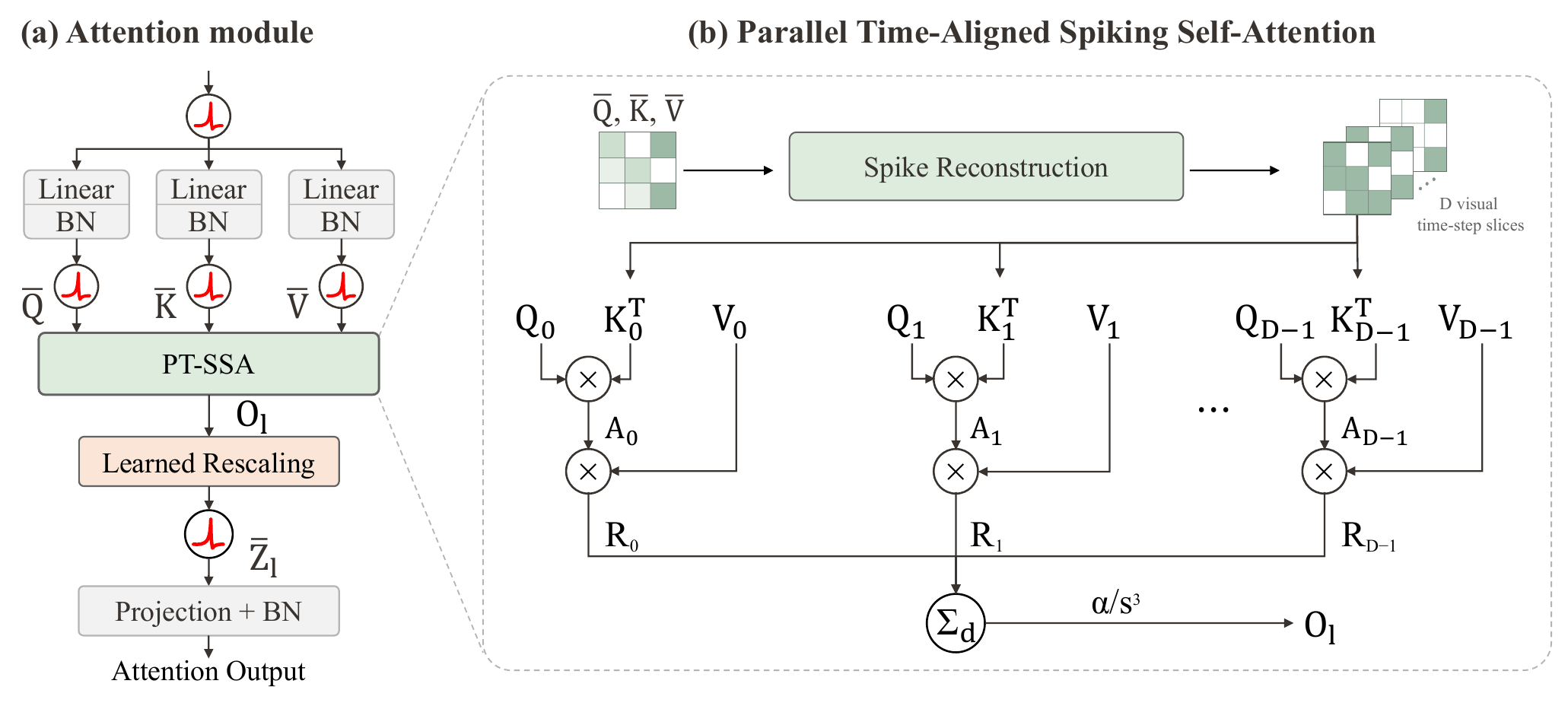}
  \caption{PT-SSA within a spiking attention block.
  (a) Compact queries, keys, and values enter PT-SSA, followed by output rescaling and firing.
  (b) Reconstructed spike slices are grouped by virtual step, and their time-aligned attention outputs are summed over the virtual time-step dimension.
  Red spike symbols indicate I-LIF or SFA spiking neurons, and $\times$ indicates matrix multiplication.
  The illustrated SFA path uses $s=D$ and a learned $\theta_l$; I-LIF uses $s=1$ and $\theta_l=1$.}
  \label{fig:ptssa-overview}
\end{figure*}

The analysis above shows that operator-level train--inference equivalence requires preserving the virtual-step alignment within $QK^{\mathsf T}V$.
PT-SSA enforces this constraint during parallel training by reconstructing the represented spike sequences and evaluating self-attention only within matching virtual steps.
Figure~\ref{fig:ptssa-overview} summarizes the resulting PT-SSA block from compact query, key, and value generation through representation-specific output firing.
The block reconstructs $D$ consecutive binary slices for each compact representation, evaluates $\mathbf{Q}_d\mathbf{K}_d^{\mathsf T}\mathbf{V}_d$ independently at each shared virtual-step index $d$, and reduces the resulting $\mathbf{R}_d$ only after attention.

\subsubsection{Spike Sequence Reconstruction}
For $X\in\{Q,K,V\}$, let $\bar{\mathbf{X}}=\mathbf{C}_{X}/s$ denote its spike-count representation and $\mathbf{C}_{X}=s\bar{\mathbf{X}}$ its firing-count tensor.
The spike tensor at virtual step $d$ is reconstructed elementwise as
\begin{equation}
  \mathbf{X}_d
  = \ind[\mathbf{C}_{X}>d]
  = \ind[s\bar{\mathbf{X}}>d].
  \label{eq:unary-expansion}
\end{equation}
For an entry with firing count $c$, Eq.~\eqref{eq:unary-expansion} produces a spike exactly when $d<c$, recovering the spike sequence defined in Eq.~\eqref{eq:consecutive-code}.
Applying this mapping to $\bar{\mathbf{Q}}$, $\bar{\mathbf{K}}$, and $\bar{\mathbf{V}}$ reconstructs the spike tensors $\{\mathbf{Q}_d,\mathbf{K}_d,\mathbf{V}_d\}_{d=0}^{D-1}$ over $D$ virtual steps.
Since each $\mathbf{X}_d$ depends only on $\bar{\mathbf{X}}$ and $d$, all $D$ spike tensors can be reconstructed in parallel.

\subsubsection{Parallel Time-Aligned Spiking Self-Attention}
At each virtual step $d$, PT-SSA computes SSA \citep{zhou2023spikformer} from the reconstructed spike tensors $\mathbf{Q}_d$, $\mathbf{K}_d$, and $\mathbf{V}_d$:
\begin{equation}
  \mathbf{A}_d=\mathbf{Q}_d\mathbf{K}_d^{\mathsf T},
  \qquad
  \mathbf{R}_d=\mathbf{A}_d\mathbf{V}_d.
  \label{eq:slice-attention}
\end{equation}
The shared index $d$ ensures that $Q$, $K$, and $V$ spikes interact only within the same virtual step, excluding every cross-time product $\mathbf{Q}_i\mathbf{K}_j^{\mathsf T}\mathbf{V}_k$ with unequal indices.
Because the $D$ outputs are independent, PT-SSA computes them in parallel, sums them, and applies the factor $\alpha/s^3$:
\begin{equation}
  \mathbf{O}_{\mathrm{PT\text{-}SSA}}
  =\frac{\alpha}{s^3}\sum_{d=0}^{D-1}\mathbf{R}_d.
  \label{eq:ptssa}
\end{equation}

\subsubsection{Time-Alignment Property}
During spike-driven inference, the count tensors $\mathbf{C}_Q$, $\mathbf{C}_K$, and $\mathbf{C}_V$ are converted into the $D$-step spike sequences defined in Eq.~\eqref{eq:consecutive-code}.
Equation~\eqref{eq:unary-expansion} reconstructs the same spike sequences from the training representations $\bar{\mathbf{Q}}$, $\bar{\mathbf{K}}$, and $\bar{\mathbf{V}}$.
Therefore, at every virtual step $d$, PT-SSA and spike-driven inference use the same $\mathbf{Q}_d$, $\mathbf{K}_d$, and $\mathbf{V}_d$ and produce the same attention output $\mathbf{R}_d$.
Both paths then sum the same $D$ outputs and apply the factor $\alpha/s^3$, yielding identical attention outputs before the subsequent firing function:
\begin{equation}
  \mathbf{O}_{\mathrm{PT\text{-}SSA}}=\mathbf{O}_{\mathrm{inference}}.
  \label{eq:alignment-property}
\end{equation}
Under the corresponding I-LIF/SFA conversion assumptions and rules \citep{luo2024integer,yao2025scaling}, Eq.~\eqref{eq:alignment-property} extends local train--inference equivalence through the subsequent spiking neuron and linear projection.

\subsubsection{Parallel Surrogate-Gradient Routing}
At each virtual step $d$, we compute the gradients of the reconstructed spike tensors $\mathbf{Q}_d$, $\mathbf{K}_d$, and $\mathbf{V}_d$ independently.
Equation~\eqref{eq:ptssa} first gives the gradient propagated to $\mathbf{R}_d$ as
\begin{equation}
  \frac{\partial\mathcal{L}}{\partial\mathbf{R}_d}
  =\frac{\alpha}{s^3}
  \frac{\partial\mathcal{L}}{\partial\mathbf{O}_{\mathrm{PT\text{-}SSA}}},
  \qquad d=0,\ldots,D-1.
  \label{eq:ptssa-gradient-routing}
\end{equation}
Applying the chain rule to the attention operation at step $d$ yields
\begin{align}
  \frac{\partial\mathcal{L}}{\partial\mathbf{V}_d}
    &=\mathbf{A}_d^{\mathsf T}
      \frac{\partial\mathcal{L}}{\partial\mathbf{R}_d}, &
  \frac{\partial\mathcal{L}}{\partial\mathbf{A}_d}
    &=\frac{\partial\mathcal{L}}{\partial\mathbf{R}_d}
      \mathbf{V}_d^{\mathsf T}, \nonumber\\
  \frac{\partial\mathcal{L}}{\partial\mathbf{Q}_d}
    &=\frac{\partial\mathcal{L}}{\partial\mathbf{A}_d}\mathbf{K}_d, &
  \frac{\partial\mathcal{L}}{\partial\mathbf{K}_d}
    &=\left(\frac{\partial\mathcal{L}}{\partial\mathbf{A}_d}\right)^{\mathsf T}\mathbf{Q}_d.
  \label{eq:ptssa-backward}
\end{align}
To propagate gradients through the non-differentiable reconstruction in Eq.~\eqref{eq:unary-expansion}, we apply a unit straight-through estimator (STE) \citep{neftci2019surrogate} with respect to the firing count $\mathbf{C}_X=s\bar{\mathbf{X}}$:
\begin{equation}
  \frac{\partial X_{d,m}}{\partial C_{X,m}}\approx 1,
  \qquad X\in\{Q,K,V\},
  \label{eq:unary-ste}
\end{equation}
where $m$ indexes an element of the corresponding tensor.
Since $\mathbf{C}_X=s\bar{\mathbf{X}}$, the corresponding surrogate derivative with respect to the compact activation is
\begin{equation}
  \frac{\partial X_{d,m}}{\partial\bar{X}_m}\approx s.
  \label{eq:unary-ste-compact}
\end{equation}
The gradients of the $D$ reconstructed spike tensors are therefore aggregated as
\begin{equation}
  \frac{\partial\mathcal{L}}{\partial\bar{\mathbf{X}}}
  \approx s\sum_{d=0}^{D-1}
  \frac{\partial\mathcal{L}}{\partial\mathbf{X}_d},
  \qquad X\in\{Q,K,V\}.
  \label{eq:unary-ste-aggregation}
\end{equation}
Equivalently, our implementation uses normalized slices $\widetilde{\mathbf{X}}_d=\mathbf{X}_d/s$ and applies the unit surrogate derivative $\partial\widetilde{X}_{d,m}/\partial\bar{X}_m\approx1$.
This implementation gives the same forward output and surrogate gradient as Eqs.~\eqref{eq:ptssa} and~\eqref{eq:unary-ste-aggregation}.
Equation~\eqref{eq:unary-ste-aggregation} allows the $D$ virtual-step gradients to be computed in parallel and then reduced to the gradient of $\bar{\mathbf{X}}$.

\subsection{Adaptive Rescaling for SFA}
\label{sec:adaptive-threshold}

PT-SSA removes the nonnegative cross-time terms in Eq.~\eqref{eq:cross-time-error}, reducing the membrane potential received by the subsequent SFA neuron for fixed input activations.
With the default unit firing threshold, the neuron may consequently emit only low firing counts and underutilize the available firing-count levels.
To accommodate block-specific output scales with negligible parameter overhead, we insert a learnable rescaling module immediately before the attention-output SFA neuron of each PT-SSA block $l$.
This module divides the attention output by one positive parameter $\theta_l$ shared within the block, adding only $L$ scalar parameters to a network with $L$ such blocks.
The SFA neuron retains its standard unit-threshold firing rule; the preceding rescaling is mathematically equivalent to adapting its effective firing threshold relative to the unscaled attention output.
We refer to PT-SSA equipped with this rescaling as Adaptive PT-SSA, and use adaptive threshold to describe the equivalent firing response.

\subsubsection{Rescaling and Firing}
Let $\mathbf{O}_{\mathrm{PT\text{-}SSA},l}$ denote the attention output after the head scaling in Eq.~\eqref{eq:ptssa}.
The learned rescaling and the subsequent standard SFA firing rule are
\begin{equation}
  \mathbf{U}_l=\theta_l^{-1}\mathbf{O}_{\mathrm{PT\text{-}SSA},l},
  \qquad
  \bar{\mathbf{Z}}_l
    =\frac{1}{D}\left\lfloor
      \clip(\mathbf{U}_l,0,D)\right\rceil.
  \label{eq:adaptive-firing}
\end{equation}
Here, $\mathbf{U}_l$ is the rescaled input to the SFA neuron, $\bar{\mathbf{Z}}_l$ is the firing-rate output supplied to the output projection, and $\theta_l=1$ makes the rescaling an identity operation.
During inference, PT-SSA can directly accumulate $\sum_d\mathbf{R}_d$ and use the equivalent firing threshold $s^3\theta_l/\alpha$, eliminating explicit scaling and division.

\subsubsection{Learning the Rescaling Parameter}
\begin{figure}[t]
  \centering
  \includegraphics[width=\columnwidth]{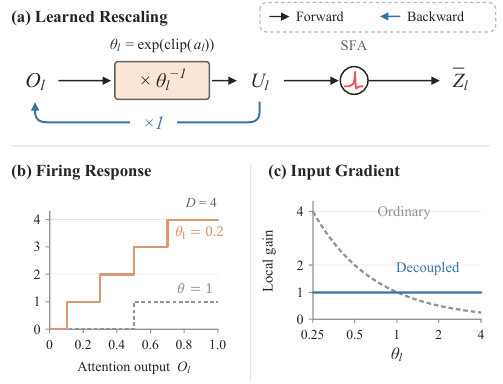}
  \caption{Adaptive rescaling and gradient routing for SFA-based PT-SSA.
  (a) Forward rescaling uses a learned $\theta_l$, while an identity STE passes the gradient from $\mathbf{U}_l$ to the attention output unchanged.
  (b) Quantized firing count versus unscaled attention output at $D=4$ for $\theta_l=1$ and $0.2$; these curves show rounded SFA responses rather than temporal IF threshold crossings.
  (c) Local input-gradient gain versus $\theta_l$ for ordinary ($1/\theta_l$) and decoupled ($1$) differentiation.}
  \label{fig:rescaling-gradients}
\end{figure}

To ensure $\theta_l>0$, we define $\theta_l=\exp\bigl(\clip(a_l,-8,8)\bigr)$ and treat $a_l$ as a learnable parameter initialized to zero.
The clipping operation bounds $\theta_l$ in the forward pass and is treated as the identity in the backward pass.
Figure~\ref{fig:rescaling-gradients} illustrates the learned forward rescaling, the resulting SFA firing response, and the decoupled gradient propagated to the attention output.
Let $M_l$ denote the number of elements in $\mathbf{O}_{\mathrm{PT\text{-}SSA},l}$.
From $U_{l,m}=O_{\mathrm{PT\text{-}SSA},l,m}/\theta_l$, we define the scaled gradient of $a_l$ as
\begin{align}
  \frac{\partial\mathcal{L}}{\partial a_l}
  \mathrel{:=}
  &\frac{1}{\sqrt{M_lD}}
   \sum_{m=1}^{M_l}
   \frac{\partial\mathcal{L}}{\partial U_{l,m}}
   \frac{\partial U_{l,m}}{\partial a_l} \nonumber\\*
  ={}&-\frac{1}{\sqrt{M_lD}}
   \sum_{m=1}^{M_l}
   \frac{\partial\mathcal{L}}{\partial U_{l,m}}
   \frac{O_{\mathrm{PT\text{-}SSA},l,m}}{\theta_l}.
  \label{eq:threshold-gradient}
\end{align}
Following LSQ, we scale the shared log-parameter gradient by $1/\sqrt{M_lD}$ to reduce its dependence on the tensor size and the firing-count range \citep{esser2020lsq}.
For the gradient propagated to PT-SSA, we use an identity STE on the attention-output branch to omit the $1/\theta_l$ factor:
\begin{equation}
  \frac{\partial\mathcal{L}}{\partial\mathbf{O}_{\mathrm{PT\text{-}SSA},l}}
  \mathrel{:=}
  \frac{\partial\mathcal{L}}{\partial\mathbf{U}_l}.
  \label{eq:decoupled-attention-gradient}
\end{equation}
The effect of this gradient-routing choice is evaluated in the component-wise ablation in Section~\ref{sec:gradient-routing-ablation}.

Algorithm~\ref{alg:adaptive-ptssa} summarizes the training forward pass and the surrogate backward rules used by Adaptive PT-SSA.

\begin{algorithm}[t]
  \caption{Adaptive PT-SSA training forward and surrogate backward}
  \label{alg:adaptive-ptssa}
  \begin{algorithmic}[1]
    \Require SFA activations $\bar{\mathbf{Q}},\bar{\mathbf{K}},\bar{\mathbf{V}}$; $D,\alpha$; log-threshold $a_l$
    \Ensure Output firing rate $\bar{\mathbf{Z}}_l$ and surrogate gradients
    \Statex \textbf{Forward}
    \State $s\gets D$ \Comment{SFA normalization}
    \State $\mathbf{C}_X \gets s\bar{\mathbf{X}}$ for $X\in\{Q,K,V\}$
    \ForAll{$d\in\{0,\ldots,D-1\}$ \textbf{in parallel}}
      \State $\widetilde{\mathbf{X}}_d \gets \ind[\mathbf{C}_X>d]/s$ for $X\in\{Q,K,V\}$
      \State $\widetilde{\mathbf{A}}_d \gets \widetilde{\mathbf{Q}}_d\widetilde{\mathbf{K}}_d^{\mathsf T}$
      \State $\widetilde{\mathbf{R}}_d \gets \widetilde{\mathbf{A}}_d\widetilde{\mathbf{V}}_d$
    \EndFor
    \State $\mathbf{O}_l \gets \alpha\sum_{d=0}^{D-1}\widetilde{\mathbf{R}}_d$
    \State $\theta_l \gets \exp\!\left(\clip(a_l,-8,8)\right)$ and $\mathbf{U}_l \gets \mathbf{O}_l/\theta_l$
    \State $\bar{\mathbf{Z}}_l \gets D^{-1}\left\lfloor\clip(\mathbf{U}_l,0,D)\right\rceil$
    \Statex \textbf{Surrogate backward}
    \State Let $\mathbf{G}_Y=\partial\mathcal{L}/\partial\mathbf{Y}$ for every tensor $\mathbf{Y}$
    \State Receive $\mathbf{G}_{\bar Z_l}$ from the subsequent network
    \State $\mathbf{G}_{U_l}\gets D^{-1}\mathbf{G}_{\bar Z_l}\odot\ind[0\leq\mathbf{U}_l\leq D]$ \Comment{SFA surrogate}
    \State $\mathbf{G}_{O_l}\gets\mathbf{G}_{U_l}$ \Comment{decoupled input gradient}
    \State $\displaystyle G_{a_l}\gets-\frac{1}{\sqrt{M_lD}}\sum_m G_{U_l,m}\frac{O_{l,m}}{\theta_l}$
    \ForAll{$d\in\{0,\ldots,D-1\}$ \textbf{in parallel}}
      \State $\mathbf{G}_{\widetilde R_d}\gets\alpha\mathbf{G}_{O_l}$
      \State $\mathbf{G}_{\widetilde V_d}\gets\widetilde{\mathbf{A}}_d^{\mathsf T}\mathbf{G}_{\widetilde R_d}$
      \State $\mathbf{G}_{\widetilde A_d}\gets\mathbf{G}_{\widetilde R_d}\widetilde{\mathbf{V}}_d^{\mathsf T}$
      \State $\mathbf{G}_{\widetilde Q_d}\gets\mathbf{G}_{\widetilde A_d}\widetilde{\mathbf{K}}_d$
      \State $\mathbf{G}_{\widetilde K_d}\gets\mathbf{G}_{\widetilde A_d}^{\mathsf T}\widetilde{\mathbf{Q}}_d$
    \EndFor
    \State $\mathbf{G}_{\bar X}\gets\sum_{d=0}^{D-1}\mathbf{G}_{\widetilde X_d}$ for $X\in\{Q,K,V\}$ \Comment{unit STE}
    \State \Return $\bar{\mathbf{Z}}_l,\mathbf{G}_{\bar Q},\mathbf{G}_{\bar K},\mathbf{G}_{\bar V},G_{a_l}$
  \end{algorithmic}
\end{algorithm}

\subsection{Efficient Implementation}
\label{sec:efficient-implementation}

Following Eq.~\eqref{eq:ptssa}, PT-SSA can be implemented in PyTorch by reconstructing $\{\mathbf{X}_d\}_{d=0}^{D-1}$ for $X\in\{Q,K,V\}$ and computing the $D$ time-aligned attention operations in parallel.
However, storing these reconstructed tensors and the corresponding attention intermediates increases activation memory and memory traffic during training.
To reduce this training overhead, we further implement a fused kernel in Triton \citep{tillet2019triton}, denoted as PT-SSA (Triton).
Within each tile, PT-SSA (Triton) reconstructs $\mathbf{Q}_d$, $\mathbf{K}_d$, and $\mathbf{V}_d$ for the current virtual step, computes $\mathbf{R}_d$, and immediately accumulates the result.
This procedure avoids materializing all reconstructed tensors and attention intermediates simultaneously.
During backpropagation, it regenerates the required $\mathbf{Q}_d$, $\mathbf{K}_d$, $\mathbf{V}_d$, and attention intermediates instead of storing them in the forward pass, trading recomputation for lower training memory.

\section{Experiments}
\label{sec:experiments}
\setcounter{dblbotnumber}{0}

We evaluate PT-SSA in terms of operator-level train--inference alignment, classification accuracy, sensitivity to the virtual expansion length, and training efficiency.
We further examine how the adaptive firing threshold changes the firing response of the output SFA neurons.

\subsection{Experimental Setup}
\label{sec:experimental-setup}

\subsubsection{Datasets and Backbones}
We evaluate static-image classification on CIFAR-10, CIFAR-100 \citep{krizhevsky2009learning}, and ImageNet-1K \citep{deng2009imagenet}.
CIFAR-10/100 contain 50,000 training and 10,000 test images at $32\times32$ resolution; ImageNet-1K contains approximately 1.28 million training and 50,000 validation images from 1,000 classes.
We denote an E-SpikeFormer backbone with $L$ attention blocks and an attention embedding dimension of $C$ as E-SpikeFormer-$L$-$C$.
Following the Spike-Driven Transformer V3 design \citep{yao2025scaling}, our CIFAR-adapted E-SpikeFormer-4-384 uses three stages with widths $96$, $192$, and $384$, followed by four attention blocks at width $384$.
We use 12 heads on CIFAR-10 and eight heads on CIFAR-100.
ImageNet experiments use the 5.11M-parameter E-SpikeFormer-8-192 backbone with stage widths $32$, $64$, $128$, and $192$ and eight attention blocks.
Within each controlled comparison, SSA and PT-SSA share the backbone topology, projections, convolutional blocks, residual paths, classifier, and attention hyperparameters.

\subsubsection{Training Protocol}
We train all models from scratch using dataset-specific optimization schedules.
We follow the dataset-specific training recipes of Spikingformer \citep{zhou2023spikingformer} for CIFAR-10/100 and Spike-Driven Transformer V3 \citep{yao2025scaling} for ImageNet-1K.
CIFAR models are optimized with AdamW for 400 cosine-schedule epochs followed by 10 cooldown epochs, using 20 warmup epochs, batch size 64, initial learning rate $5\times10^{-4}$, weight decay 0.06, and label smoothing 0.1.
We use random horizontal flipping, RandAugment, mixup with strength 0.5, and random erasing with probability 0.25.
Mixup is disabled after the first 200 epochs.
ImageNet models are trained for 200 epochs on eight GPUs with LAMB, an effective batch size of 2,048, five warmup epochs, weight decay 0.05, and stochastic-depth rate 0.1.
ImageNet SSA and PT-SSA models use the same learning rate of $4.8\times10^{-3}$.
ImageNet inputs are randomly augmented and erased during training and center-cropped to $224\times224$ for validation.
Training and evaluation use PyTorch \citep{paszke2019pytorch} and automatic mixed precision.
For PT-SSA reconstruction, we use a unit STE with respect to the firing count, equivalently implemented using normalized spike slices with a unit surrogate derivative.

\subsubsection{Representations and Evaluation}
Following Eq.~\eqref{eq:unified-representation}, we set $s=1$ for I-LIF count activations in $\{0,\ldots,D\}$ and $s=D$ for SFA firing-rate activations in $\{0,1/D,\ldots,1\}$ \citep{yao2025scaling}.
Integer-valued and spike-driven inference are evaluated from the same checkpoint.
At spike-driven inference, each count $C$ is expanded into the spike sequence $S_d(C)=\ind[C>d]$, and attention is evaluated over $D$ virtual steps.
For $n$ paired evaluation runs, we define the reported mean absolute Top-1 gap as
\begin{equation}
  |\Delta|
  =\frac{1}{n}\sum_{r=1}^{n}
  \left|\mathrm{Acc}_{\mathrm{int},r}-\mathrm{Acc}_{\mathrm{spk},r}\right|.
  \label{eq:accuracy-gap}
\end{equation}
The mean accuracies of the two inference paths are reported separately, so their difference need not equal the paired mean absolute gap in Eq.~\eqref{eq:accuracy-gap}.
For CIFAR-10 and CIFAR-100, we report the mean and sample standard deviation of accuracies and per-run absolute gaps over two random seeds.
For ImageNet-1K, we use a single random seed because of the high training cost.
For the paired re-evaluation, AMP FP16 and TF32 are enabled, while cuDNN algorithm benchmarking is disabled and deterministic algorithms are selected.

\begin{table*}[t]
  \centering
  \caption{Train--inference alignment for I-LIF counts and SFA firing rates at $D=4$.
  Integer-valued and spike-driven results in each row come from the same checkpoint.
  The paired absolute Top-1 gap $|\Delta|$ is reported in percentage points and computed before rounding.}
  \label{tab:main-results}
  \small
  \resizebox{\textwidth}{!}{%
  \begin{tabular}{@{}lllcccc@{}}
    \toprule
    Dataset & Backbone & Representation & Attention path & Integer Top-1 (\%) $\uparrow$ & Spike-driven Top-1 (\%) $\uparrow$ & $|\Delta|$ (pp) $\downarrow$ \\
    \midrule
    \multirow{4}{*}{CIFAR-10} & \multirow{4}{*}{E-SpikeFormer-4-384} & I-LIF count & SSA & $\mathbf{97.26\pm0.06}$ & $96.85\pm0.33$ & $0.41\pm0.27$ \\
    & & I-LIF count & PT-SSA & $97.20\pm0.07$ & $\mathbf{96.95\pm0.11}$ & $\mathbf{0.25\pm0.04}$ \\
    & & SFA rate & SSA & $97.08\pm0.19$ & $91.50\pm0.00$ & $5.58\pm0.19$ \\
    & & SFA rate & Adaptive PT-SSA & $\mathbf{97.23\pm0.03}$ & $\mathbf{97.23\pm0.09}$ & $\mathbf{0.05\pm0.01}$ \\
    \midrule
    \multirow{4}{*}{CIFAR-100} & \multirow{4}{*}{E-SpikeFormer-4-384} & I-LIF count & SSA & $\mathbf{80.69\pm0.11}$ & $78.84\pm0.03$ & $1.85\pm0.08$ \\
    & & I-LIF count & PT-SSA & $80.39\pm0.14$ & $\mathbf{80.10\pm0.18}$ & $\mathbf{0.29\pm0.04}$ \\
    & & SFA rate & SSA & $80.82\pm1.10$ & $53.83\pm1.53$ & $26.99\pm0.42$ \\
    & & SFA rate & Adaptive PT-SSA & $\mathbf{82.35\pm0.42}$ & $\mathbf{82.42\pm0.46}$ & $\mathbf{0.07\pm0.04}$ \\
    \midrule
    \multirow{2}{*}{ImageNet-1K} & \multirow{2}{*}{E-SpikeFormer-8-192} & SFA rate & SSA & \textbf{75.41} & 47.62 & 27.78 \\
    & & SFA rate & Adaptive PT-SSA & 74.59 & \textbf{74.53} & \textbf{0.06} \\
    \bottomrule
  \end{tabular}%
  }
\end{table*}

\begin{table*}[t]
  \centering
  \caption{Training efficiency on CIFAR-100 Spikingformer using one RTX 5090 ($B=64$, $D=4$, AMP FP16).
  SSA uses Torch, while recurrent LIF neurons use the SpikingJelly CuPy backend.
  Timings are averaged over 50 synchronized iterations after 20 warmup iterations and include forward and backward computation; data loading, optimizer updates, and Triton compilation are excluded.
  Peak memory excludes optimizer state; bold and underlined values indicate the best and second-best results.}
  \label{tab:triton-ablation}
  \small
  \resizebox{\textwidth}{!}{%
  \begin{tabular}{@{}llcccc@{}}
    \toprule
    Neuron & Attention implementation & Peak memory (GiB) $\downarrow$ & Forward (ms) $\downarrow$ & Backward (ms) $\downarrow$ & Throughput (sample/s) $\uparrow$ \\
    \midrule
    LIF & SSA & 4.193 & 12.909 & 19.197 & 1993.4 \\
    \midrule
    I-LIF & SSA & \underline{0.679} & \textbf{4.099} & \textbf{6.348} & \textbf{6126.5} \\
    I-LIF & PT-SSA (Torch) & 0.761 & 5.211 & 7.517 & 5028.3 \\
    I-LIF & PT-SSA (Triton) & \textbf{0.666} & \underline{4.522} & \underline{6.518} & \underline{5797.1} \\
    \bottomrule
  \end{tabular}%
  }
\end{table*}

\subsection{Operator-Level TIM}
\label{sec:ilif-equivalence}

We first perform an operator-level consistency check before measuring end-to-end classification accuracy.
For each attention block, we construct the spike-driven reference $\frac{\alpha}{s^3}\sum_{d=0}^{D-1}(Q_dK_d^{\mathsf T})V_d$ from the integer-valued $Q$, $K$, and $V$ activations.
The comparison is performed at the membrane potential supplied to the attention-output neuron, corresponding to Eq.~\eqref{eq:alignment-property}.
For each method, the spike-driven reference is constructed from its own quantized $Q$, $K$, and $V$; the SSA baseline applies attention to their compressed activations, whereas PT-SSA computes time-aligned attention.
This check therefore isolates differences in attention computation from discrepancies introduced by preceding or subsequent layers.
Figure~\ref{fig:alignment-summary}(a) summarizes this operator-level consistency check.
At $D=4$, replacing SSA with PT-SSA reduces the relative $L_2$ error from 0.7136 to $2.11\times10^{-4}$ for I-LIF and from 7.4120 to $1.60\times10^{-4}$ for SFA, while the cosine similarity rounds to 1.000 in both cases.
For SFA, this comparison isolates TIM in SDT-V3's E-SDSA and shows that PT-SSA removes it without relying on changes elsewhere in the network.

\subsection{Main Results on CIFAR and ImageNet}
\label{sec:main-results}

\begin{figure}[t]
  \centering
  \includegraphics[width=\columnwidth]{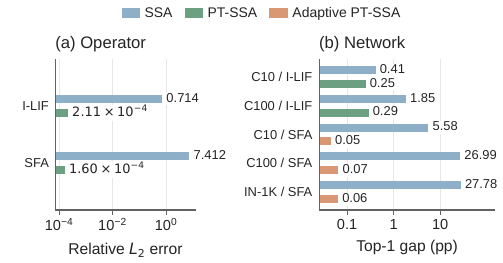}
  \caption{Operator-level and end-to-end train--inference alignment at $D=4$ (logarithmic horizontal axes; lower is better).
  (a) Relative $L_2$ error against the aligned spike reference on CIFAR-100 (I-LIF) and ImageNet-1K (SFA).
  (b) Absolute Top-1 gap in percentage points on CIFAR-10 (C10), CIFAR-100 (C100), and ImageNet-1K (IN-1K).}
  \label{fig:alignment-summary}
\end{figure}

We evaluate classification performance through controlled comparisons between SSA and PT-SSA under the same backbone and training protocol.
Table~\ref{tab:main-results} compares SSA with PT-SSA and Adaptive PT-SSA under these controlled settings.
Both integer-valued inference accuracy and spike-driven inference accuracy are reported to measure the train--inference gap directly.
Figure~\ref{fig:alignment-summary}(b) summarizes the resulting gap across datasets and activation representations.

\begin{table*}[t]
  \centering
  \caption{CIFAR-100 accuracy and train--inference gap across virtual expansion lengths.
  Values are means $\pm$ sample standard deviations over two runs.}
  \label{tab:representation-time-ablation}
  \small
  \begin{tabular}{@{}llcccc@{}}
    \toprule
    Representation & Attention Impl. & $D$ & Integer Top-1 (\%) $\uparrow$ & Spike-driven Top-1 (\%) $\uparrow$ & $|\Delta|$ (pp) $\downarrow$ \\
    \midrule
    \multirow{8}{*}{I-LIF count} & \multirow{4}{*}{SSA} & 1 & $77.31\pm0.42$ & $77.31\pm0.42$ & $0.00\pm0.00$ \\
    & & 2 & $79.88\pm0.16$ & $79.26\pm0.04$ & $0.62\pm0.12$ \\
    & & 4 & $80.69\pm0.11$ & $78.84\pm0.03$ & $1.85\pm0.08$ \\
    & & 8 & $80.90\pm0.14$ & $76.57\pm0.39$ & $4.34\pm0.53$ \\
    \cmidrule(l){2-6}
    & \multirow{4}{*}{PT-SSA} & 1 & $77.31\pm0.42$ & $77.31\pm0.42$ & $0.00\pm0.00$ \\
    & & 2 & $80.12\pm0.11$ & $79.91\pm0.12$ & $0.22\pm0.01$ \\
    & & 4 & $80.39\pm0.14$ & $80.10\pm0.18$ & $0.29\pm0.04$ \\
    & & 8 & $80.09\pm0.03$ & $79.88\pm0.22$ & $0.22\pm0.19$ \\
    \midrule
    \multirow{8}{*}{SFA rate} & \multirow{4}{*}{SSA} & 1 & $73.25\pm0.24$ & $73.25\pm0.24$ & $0.00\pm0.00$ \\
    & & 2 & $78.94\pm0.09$ & $71.40\pm0.40$ & $7.54\pm0.31$ \\
    & & 4 & $80.82\pm1.10$ & $53.83\pm1.53$ & $26.99\pm0.42$ \\
    & & 8 & $80.96\pm0.56$ & $55.21\pm1.58$ & $25.75\pm2.14$ \\
    \cmidrule(l){2-6}
    & \multirow{4}{*}{Adaptive PT-SSA} & 1 & $69.36\pm0.21$ & $69.36\pm0.21$ & $0.00\pm0.00$ \\
    & & 2 & $79.07\pm0.11$ & $79.07\pm0.05$ & $0.11\pm0.00$ \\
    & & 4 & $82.35\pm0.42$ & $82.42\pm0.46$ & $0.07\pm0.04$ \\
    & & 8 & $82.39\pm0.39$ & $82.11\pm0.56$ & $0.28\pm0.17$ \\
    \bottomrule
  \end{tabular}
\end{table*}

\subsubsection{I-LIF Count Results}
PT-SSA reduces the train--inference gap under the I-LIF count representation at $D=4$ on both CIFAR datasets.
On CIFAR-10, PT-SSA reduces the mean gap from $0.41\pm0.27$ to $0.25\pm0.04$ percentage points, while both paths retain approximately 97\% Top-1 accuracy.
On CIFAR-100, PT-SSA reduces the mean gap from $1.85\pm0.08$ to $0.29\pm0.04$ percentage points and improves mean spike-driven inference accuracy from $78.84\pm0.03$\% to $80.10\pm0.18$\%.
The corresponding mean integer-valued inference accuracy decreases from $80.69\pm0.11$\% to $80.39\pm0.14$\%, indicating that the improvement arises from stronger train--inference alignment rather than a higher integer-valued inference accuracy.

\subsubsection{SFA Results}
TIM produces substantially larger train--inference gaps under the SFA firing-rate representation.
On CIFAR-10, Adaptive PT-SSA reduces the mean gap from 5.58 to 0.05 percentage points and reaches $97.23\pm0.09$\% spike-driven inference accuracy.
On CIFAR-100, it improves spike-driven inference accuracy from $53.83\pm1.53$\% to $82.42\pm0.46$\% and reduces the mean gap from 26.99 to 0.07 percentage points.
Table~\ref{tab:adaptive-threshold-analysis} further isolates the effect of the adaptive firing threshold at $D=4$.

ImageNet-1K confirms the same behavior at a larger scale.
SSA attains 75.41\% integer-valued accuracy but only 47.62\% spike-driven inference accuracy, corresponding to a 27.78-percentage-point gap.
Under identical attention hyperparameters, Adaptive PT-SSA reaches 74.59\% and 74.53\% on the two paths, respectively, reducing the gap to 0.06 percentage points.
Its integer-valued accuracy remains 0.81 percentage points below SSA, showing that the recovered inference accuracy is not caused by a stronger integer-valued model.
The corresponding integer-valued and spike-driven Top-5 accuracies are 92.04\% and 92.01\%, respectively.

\subsection{Training Efficiency}
\label{sec:triton-ablation}

We compare the training efficiency of recurrent LIF SSA, I-LIF SSA, and the Torch and Triton implementations of PT-SSA.
All methods use the same CIFAR-100 Spikingformer tensor shapes and are evaluated on one RTX 5090.

Table~\ref{tab:triton-ablation} shows that PT-SSA (Torch) and PT-SSA (Triton) achieve $2.52\times$ and $2.91\times$ the throughput of recurrent LIF SSA, respectively, while reducing peak memory by 81.9\% and 84.1\%.
Materializing the reconstructed spike tensors introduces additional overhead in PT-SSA (Torch).
The Triton implementation reduces this overhead, improving throughput by 15.3\% and reducing peak memory by 12.5\% relative to PT-SSA (Torch).
With these gains, PT-SSA (Triton) achieves throughput only 5.4\% below that of I-LIF SSA.

\subsection{Ablation Studies}
\label{sec:ablations}

We conduct all ablation studies on CIFAR-100 using the backbone and training protocol described above.

\subsubsection{Representation and Virtual Expansion Length}
\label{sec:representation-time-ablation}

\begin{figure}[t]
  \centering
  \includegraphics[width=\columnwidth]{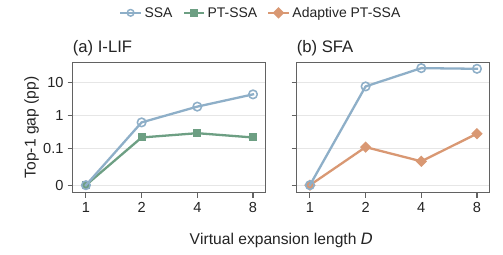}
  \caption{Mean absolute Top-1 train--inference gap versus virtual expansion length $D$ on CIFAR-100.
  Panels (a) and (b) show I-LIF and SFA, respectively; their vertical axes report percentage points on the same symmetric-log scale with a linear region around zero.}
  \label{fig:virtual-length}
\end{figure}

We examine the effect of the virtual expansion length $D$ on TIM and the adaptive firing threshold.
As $D$ increases, the SSA training path contains more off-diagonal triplets in Eq.~\eqref{eq:cross-time-error}, whereas PT-SSA retains only the $D$ time-aligned triplets.
Under SFA, changing $D$ also changes the membrane-potential scale relative to the firing threshold of the attention-output neuron.
The SFA comparison evaluates how TIM changes with $D$ under SSA and whether Adaptive PT-SSA maintains train--inference alignment.
Figure~\ref{fig:virtual-length} visualizes the gap trends, while Table~\ref{tab:representation-time-ablation} reports the corresponding integer-valued and spike-driven accuracies.

\begin{table*}[t]
  \centering
  \caption{Adaptive-threshold and gradient-routing analysis for SFA-based PT-SSA on CIFAR-100 ($D=4$).
  Top-1 and gap entries are means $\pm$ sample standard deviations over two runs.
  Firing rates and entropies are computed on the complete test set for each block and run, then averaged; $\theta_l$ statistics summarize four blocks and two runs.}
  \label{tab:adaptive-threshold-analysis}
  \small
  \resizebox{\textwidth}{!}{%
  \begin{tabular}{@{}lcccccc@{}}
    \toprule
    Setting & $\theta_l$ (mean $\pm$ std) & Firing rate (\%) & Entropy (bits) $\uparrow$ & Integer Top-1 (\%) $\uparrow$ & Spike-driven Top-1 (\%) $\uparrow$ & $|\Delta|$ (pp) $\downarrow$ \\
    \midrule
    Fixed unit & $1.00\pm0.00$ & 10.94 & 1.204 & $80.78\pm0.09$ & $81.03\pm0.01$ & $0.25\pm0.10$ \\
    Fixed $1/D^3$ & $0.02\pm0.00$ & 97.73 & 0.367 & $80.48\pm0.40$ & $80.61\pm0.40$ & $0.13\pm0.01$ \\
    Adaptive (coupled) & $0.21\pm0.02$ & 16.67 & 1.487 & $81.05\pm0.46$ & $81.46\pm0.18$ & $0.42\pm0.28$ \\
    Adaptive (decoupled) & $0.20\pm0.03$ & 17.97 & \textbf{1.519} & $\mathbf{82.35\pm0.42}$ & $\mathbf{82.42\pm0.46}$ & $\mathbf{0.07\pm0.04}$ \\
    \bottomrule
  \end{tabular}%
  }
\end{table*}

\subsubsection{Effect of $D$ on TIM}
At $D=1$, SSA and PT-SSA reduce to the same attention operation; under I-LIF their shared checkpoint yields identical integer-valued and spike-driven predictions.
As shown in Figure~\ref{fig:virtual-length}(a), when $D$ increases to 2, 4, and 8 under I-LIF, the mean SSA gaps are 0.62, 1.85, and 4.34 percentage points, whereas the corresponding PT-SSA gaps remain at or below 0.29 percentage points.
Consequently, at $D=8$, PT-SSA improves mean spike-driven inference accuracy by 3.31 percentage points despite having 0.81 percentage points lower mean integer-valued accuracy.
This difference between integer-valued and spike-driven rankings highlights the need to evaluate both paths.

Figure~\ref{fig:virtual-length}(b) shows that under SFA, the mean SSA gap is zero at $D=1$, reaches 7.54 percentage points at $D=2$, and exceeds 25 percentage points at $D=4$ and $D=8$.
Adaptive PT-SSA instead maintains a mean gap of at most 0.28 percentage points across all four expansion lengths, although its $D=8$ gap is larger than at $D=4$.

\subsubsection{PT-SSA Component Ablation}
At $D=4$, as reported in Table~\ref{tab:adaptive-threshold-analysis}, replacing SSA with PT-SSA under the unit threshold increases spike-driven inference accuracy from $53.83\pm1.53$\% to $81.03\pm0.01$\%.
The mean train--inference gap decreases from 26.99 to 0.25 percentage points, while integer-valued accuracy remains nearly unchanged.
This comparison attributes the primary accuracy recovery to time-aligned attention.
Learning the firing threshold on top of PT-SSA further improves mean integer-valued and spike-driven inference accuracy by 1.58 and 1.39 percentage points, respectively, as summarized in Tables~\ref{tab:representation-time-ablation} and~\ref{tab:adaptive-threshold-analysis}.

\subsubsection{Gradient Routing for the Adaptive Threshold}
\label{sec:gradient-routing-ablation}

We next isolate the effect of the gradient-routing rule while keeping the forward threshold parameterization unchanged.
Table~\ref{tab:adaptive-threshold-analysis} summarizes this comparison together with the corresponding firing-response statistics.
Both variants use per-block thresholds initialized with $\theta_l=1$, the same normalized threshold gradient, and the same nominal training protocol.
The coupled variant propagates $\partial\mathcal{L}/\partial\mathbf{O}_l=\mathbf{G}_l/\theta_l$, whereas the decoupled variant propagates $\partial\mathcal{L}/\partial\mathbf{O}_l=\mathbf{G}_l$.

Table~\ref{tab:adaptive-threshold-analysis} shows that, across two runs, decoupling improves integer-valued Top-1 accuracy from $81.05\pm0.46$\% to $82.35\pm0.42$\% and spike-driven Top-1 accuracy from $81.46\pm0.18$\% to $82.42\pm0.46$\%.
It also reduces the mean absolute train--inference gap from $0.42\pm0.28$ to $0.07\pm0.04$ percentage points.
These results support separating the learned threshold scale from the PT-SSA surrogate-gradient magnitude.
Because both variants use the same $1/\sqrt{M_lD}$ threshold-gradient normalization, this comparison does not isolate the effect of that normalization factor.

\subsection{Analysis of the Adaptive Firing Threshold}
\label{sec:adaptive-threshold-analysis}

We analyze how the learned threshold changes the firing response of SFA-based PT-SSA.
For each attention-output SFA neuron in block $l$, we measure the mean firing rate
\begin{equation}
  \rho_l
  =\frac{1}{N_l}\sum_{m=1}^{N_l}\bar Z_{l,m}
  =\frac{1}{N_lD}\sum_{m=1}^{N_l}C_{Z,l,m},
  \label{eq:mean-firing-rate}
\end{equation}
where $N_l$ is the total number of output elements collected over the complete test set in one run and $C_{Z,l,m}=D\bar Z_{l,m}$ is the quantized firing count.
This rate equals the fraction of active spikes in the corresponding virtual spike sequences.
We also compute the firing-count entropy $H_l=-\sum_{c=0}^{D}p_{l,c}\log_2 p_{l,c}$, where $p_{l,c}$ is the empirical frequency of count $c$ among those $N_l$ elements and $0\log_2 0=0$.
Both statistics are computed separately for each block and run, then averaged equally across the four attention blocks and two runs.
We compare the unit threshold, a fixed threshold of $1/D^3$, and the proposed per-block threshold learned with Eq.~\eqref{eq:threshold-gradient}.

Table~\ref{tab:adaptive-threshold-analysis} shows that the fixed $1/D^3$ threshold yields a firing rate of 97.73\% but an entropy of only 0.367 bits, indicating that outputs concentrate near the maximum firing count.
The unit threshold yields a firing rate of 10.94\% and an entropy of 1.204 bits.
The decoupled adaptive SFA neurons learn a mean threshold of 0.195, increasing the firing rate to 17.97\% and entropy to 1.519 bits while attaining the highest spike-driven inference accuracy among the compared settings.
The coupled-gradient variant has a similar firing response, with a mean threshold of 0.209, a firing rate of 16.67\%, and an entropy of 1.487 bits, but its spike-driven accuracy is lower.
These results suggest that threshold adaptation improves the utilization of the available firing-count levels; increasing the mean firing rate alone does not ensure higher accuracy.

\section{Conclusion}
\label{sec:conclusion}

This work identifies Temporal Interaction Mismatch (TIM), a structural inconsistency between compressed integer-valued training and temporal spike-driven inference in SFA-based spiking self-attention.
To address this issue, we propose PT-SSA, which reconstructs consecutive virtual spike slices and performs attention only between time-aligned query, key, and value slices, together with Adaptive PT-SSA, which learns per-block firing thresholds for SFA outputs.
Experiments on CIFAR-10, CIFAR-100, and ImageNet-1K demonstrate that the proposed methods substantially reduce the integer--spike accuracy gap, reducing the ImageNet-1K Top-1 gap from 27.78 to 0.06 percentage points while achieving 74.53\% spike-driven accuracy.
The Triton-fused PT-SSA training kernel further reduces the training overhead of aligned attention and retains a $2.91\times$ throughput advantage over recurrent LIF SSA.
By eliminating TIM while preserving spike-driven inference semantics, PT-SSA provides a promising foundation for the future deployment of integer-trained spiking Transformers on neuromorphic hardware.


\section*{Declaration of competing interest}

The authors declare that they have no known competing financial interests or personal relationships that could have appeared to influence the work reported in this paper.

\section*{Data availability}

The code and data are available upon request.

\bibliographystyle{cas-model2-names}
\bibliography{references}

\end{document}